\documentclass[journal]{IEEEtran}
\usepackage{cite}

\usepackage{xcolor}

\ifCLASSINFOpdf
  \usepackage[pdftex]{graphicx}
  \graphicspath{{../pdf/}{../jpeg/}}
  \DeclareGraphicsExtensions{.pdf,.jpeg,.png}
\else
\fi

\usepackage{enumitem}
\usepackage{newtxmath}
\usepackage{hyperref}
\usepackage{inconsolata}
\usepackage{multirow}

\begin{document}

\title{ParetoLens: A Visual Analytics Framework for Exploring Solution Sets of Multi-objective Evolutionary Algorithms}

\author{
        Yuxin~Ma, Zherui~Zhang, Ran~Cheng,
        \textit{Southern University of Science and Technology, China}

        Yaochu~Jin, \textit{Westlake University, China}
        
        Kay Chen~Tan, \textit{Hong Kong Polytechnic University, Hong Kong SAR, China}
\thanks{Corresponding author: Ran Cheng (ranchengcn@gmail.com)}
}


\markboth{Journal of \LaTeX\ Class Files,~Vol.~XX, No.~X, XXXX}%
{Shell \MakeLowercase{\textit{et al.}}: Bare Demo of IEEEtran.cls for IEEE Journals}

\maketitle

\begin{abstract}

In the domain of multi-objective optimization, evolutionary algorithms are distinguished by their capability to generate a diverse population of solutions that navigate the trade-offs inherent among competing objectives. This has catalyzed the ascension of evolutionary multi-objective optimization (EMO) as a prevalent approach. Despite the effectiveness of the EMO paradigm, the analysis of resultant solution sets presents considerable challenges. This is primarily attributed to the high-dimensional nature of the data and the constraints imposed by static visualization methods, which frequently culminate in visual clutter and impede interactive exploratory analysis. To address these challenges, this paper introduces ParetoLens, a visual analytics framework specifically tailored to enhance the inspection and exploration of solution sets derived from the multi-objective evolutionary algorithms. Utilizing a modularized, algorithm-agnostic design, ParetoLens enables a detailed inspection of solution distributions in both decision and objective spaces through a suite of interactive visual representations. This approach not only mitigates the issues associated with static visualizations but also supports a more nuanced and flexible analysis process. The usability of the framework is evaluated through case studies and expert interviews, demonstrating its potential to uncover complex patterns and facilitate a deeper understanding of multi-objective optimization solution sets. A demo website of ParetoLens is available at \texttt{\url{https://dva-lab.org/paretolens/}}.
\end{abstract}


%
\IEEEpeerreviewmaketitle

\section{Introduction}
\label{sec:introduction}

\IEEEPARstart{R}{eal}-world optimization challenges often involve multiple objectives to be considered simultaneously, categorizing them as multi-objective optimization problems (MOPs). Taking environmental management~\cite{MAYER2020115058} as an example, stakeholders typically engage in tasks that involve developing economic plans accounting for profit, environmental impact, and social fairness simultaneously. Applying conventional optimization techniques to such MOPs can be insufficient due to the complex and often conflicting dynamics among the various objectives. By contrast, the evolutionary multi-objective optimization (EMO) paradigm offers a systematic and universal approach for trading off among the multiple optimization objectives~\cite{Tian2021}. Under the paradigm, multi-objective evolutionary algorithms (MOEAs) prove to be effective in approximating the Pareto optimal set and fostering a deeper understanding of the relationships between different objectives~\cite{Li2018}.

Upon the generation of solutions derived from MOEAs, the subsequent phase necessitates an understanding of these results. To this end, visualization techniques have been demonstrated to be effective in gaining insights from the solution sets, such as identifying the data distributions of solutions or trade-offs among multiple objectives~\cite{holden2004visualization,He2016,Nagar2022}. This visual inspection process is not only critical for assessing the performance of different algorithms, but it also bolsters successive stages of multi-criteria decision-making, particularly when the application context is involved in the visualization perspectives. 

The main data type for multi-objective optimization comprises high-dimensional vectors in both decision and objective spaces, which are inherently associated with the solutions. 
Thus, a plethora of high-dimensional visualization techniques have been leveraged in prior studies \cite{Li2017,Nagar2022}, including projection-based approaches, parallel coordinates plots, and their enhanced variants. 
Notably, methods such as iSOM~\cite{Nagar2022}, 3D-RadVis~\cite{Ibrahim2016-3dradvis}, PaletteViz~\cite{Talukder2020}, and PaletteStarViz~\cite{Talukder2020star} employ advanced projection algorithms capable of translating objective vectors into scatter plots in two- or three-dimensional spaces to render them comprehensible to human observers. 
Moving beyond traditional multi-dimensional scaling (MDS) techniques, Walter \textit{et al.}~\cite{Walter2022tevc} introduced a novel algorithm that incorporates landmarks to reduce computational demands associated with generating projections. Additionally, He \textit{et al.}~\cite{He2021knee} proposed a knee-based decision-making strategy, wherein a visualization component is crucial for illustrating key features of the Pareto front.

While the static visualization methods mentioned above are effective at delivering key insights from the solution sets, they inherently lack interactivity, thereby presenting data in a fixed format. This static nature introduces certain limitations:

\begin{itemize}[leftmargin=*]
    \item First, although these visualizations are frequently used in the presentation of experimental results from MOEAs in academic literature~\cite{Li2017}, the propensity for visual clutter is a major drawback. 
    Such issue arises when visual elements of corresponding data instances, such as edges in parallel coordinates plots (PCP) and dots in scatterplots, show severe overlap, thereby significantly hampering the readability of the visualized data.

    \item Second, the analysis of solutions generated by MOEAs constitutes a complex endeavor that may not be adequately addressed through simple non-interactive visual presentations. 
    An exploratory analysis framework is deemed necessary to facilitate the identification and understanding of intricate patterns, correlations, and trends. 
    Analysts may require an iterative and multi-faceted approach to rapidly validate hypotheses, refine data filters, and adjust parameters for generating new visualizations, thereby uncovering a wealth of information across multiple dimensions and solution sets. 
    In general, the aforementioned challenges underline the imperative for an interactive analysis paradigm that augments the capabilities of static visualization techniques.
\end{itemize}

Following the development in visualization theory and practices over the past decades, visual analytics, emerged as a distinct research direction, has shown its strength in establishing a task-oriented exploratory analysis methodology, where human analysts can participate in the analytical process~\cite{keim2006challenges,keim2010mastering,Tominski20IVDA}. 
To handle large and complex datasets, such methodology leverages the capabilities of automated computational methods to distill salient features and abstractions, thereby enhancing human comprehension~\cite{Tominski20IVDA}. The interactive visualizations incorporated in visual analytics frameworks augment the flexibility and representational potency of static visualizations, which are applying in facilitating the exploration of complex datasets.

The challenges identified above illuminate that the urgent requirements for solution analysis in MOEAs align effectively with the visual analytics paradigm. In this paper, a visual analytics framework,~\textit{ParetoLens}, is proposed to facilitate inspection and exploration of solution sets from MOEAs. Predicated on a modularized and algorithm-agnostic design, ParetoLens has the capability to accommodate a diverse range of MOEAs once they satisfy the same multi-objective optimization output protocol. Analysts can scrutinize the distributions of the solution set in both decision and objective spaces through a multi-aspect visualization design. Real-time interactions are also applied to foster thorough exploration and hypothesis evaluation on potentially salient patterns in the solution set. The effectiveness of the framework is demonstrated through case studies and expert interviews on benchmarking test problems.

In summary, our contributions include: 

\begin{itemize}[leftmargin=*]
    \item A visual analytics framework, ParetoLens, for inspection and exploration of solution sets generated from MOEAs;

    \item A suite of visual representations and interactions that reveals the distributions and patterns hidden within the solution set in the decision and the objective spaces;

    \item A web-based implementation of the proposed framework that supports user-friendly access via web browsers with high usability and interactivity.
\end{itemize}

\section{Related Work}

This section reviews relevant literature on visualization in the realm of multi-objective optimization and explainable AI, both of which are closely related to the research topic under discussion.

\subsection{Visualization in Evolutionary Multi-objective Optimization}
\label{sec:related-work-emo-vis}
Owing to the opaque nature of multi-objective optimization procedures, comprehending and elucidating the internal workings of the algorithms pose continual challenges for human analysts~\cite{He2016}. Thus, various visualization approaches have been developed to open the black-box of these processes and offer insights into the solutions. Following the taxonomy outlined by Filipi\u{c} and Tu\u{s}ar~\cite{Filipic2018}, the methods for presenting solutions fall into two categories: visualizing single or multiple solution sets. This paper focuses on reviewing visualization literature for individual solution sets, which is the most relevant to our work.

\subsubsection{Projection-based Methods} Dimensionality reduction, which involves mapping data into visually perceptible spaces (typically 2-D or 3-D), represents a fundamental strategy in visualizing high-dimensional data. This strategy mirrors the characteristics involved in analyzing potentially high-dimensional solution sets. Widely-adopted multi-dimensional projection techniques, such as PCA~\cite{pcabook2014}, t-SNE ~\cite{van2008visualizing}, and UMAP~\cite{mcinnes2018umap}, have been extensively utilized across different scenarios. In recent years, the MOEA literature has started to introduce and deliberate on novel projection techniques specifically designed for solution set visualization. For instance, the objective reduction-based visualization method (ORV) proposed by Zhen \textit{et al.}~\cite{Zhen2020} aims to generate low-dimensional representations of solution sets while preserving distributional and Pareto-dominance information among solutions. The iSOM method~\cite{Nagar2022,Yadav2023} leverages the concept of self-organizing maps to visualize the trade-off solutions. Building upon the original 2-D RadViz design~\cite{hoffman1997dna,daniels2012properties}, 3D-RadViz~\cite{Ibrahim2016-3dradvis} extends this concept by placing anchor points and scatters in a three-dimensional visual space. PaletteViz~\cite{Talukder2020} employs a different extension methodology by segmenting solutions into multiple layers according to their distances from a selected reference point. A further variation, PaletteStarViz~\cite{Talukder2020star}, replaces RadViz with star coordinates \cite{Kandogan2000, Lehmann2013} to avoid the issues of nonlinear distortions and low flexibility of anchor points inherent in RadViz~\cite{Rubio-Sanchez2015}.

\subsubsection{Axis-based Methods} In addition to visualizing solutions in the scatterplot-based format, axis-based visualization techniques are also employed to represent distributions and salient patterns. As an enhancement to the traditional 2-D scatterplots which illustrate relationships between two variables, scatterplot matrices (SPLOM)~\cite{scatterplotbook1987,yuan2013splom} enable the simultaneous presentation of multiple variables by arranging scatterplots in an array-like manner. Another prevalent method for reconfiguring the axis layout is the parallel coordinates plot (PCP)~\cite{Inselberg2009}, where all axes are aligned in parallel. This arrangement facilitates the disclosure of multivariate patterns based on the polyline edges between two or across multiple axes~\cite{Liu2017}. To evaluate the suitability of PCPs for visualizing multi-objective solutions, Li \textit{et al.}~\cite{Li2017} discussed how edge patterns in PCPs align with data distributions or trade-offs between solutions. Observations reported in the literature suggest that PCPs can serve as an auxiliary tool for comprehending key characteristics of solution sets, such as convergence, coverage, and uniformity. However, additional validation methods such as quality metrics should be simultaneously employed. In addition to parallel-placed axes, He \textit{et al.}~\cite{He2016} proposed a radial plot to visualize the shape and location of the Pareto front as well as the solution distribution. The convexity of the front can be depicted through different line patterns on the 2-D plane.

\subsubsection{Discussion}
While there already exist a plethora of methods for presenting solutions derived from MOEAs, static visualization methods, as discussed in Section~\ref{sec:introduction}, remain insufficient for supporting interactive explorations of solution sets. Furthermore, many conventional visualization designs show difficulties when managing a large volume of data items, primarily due to visual clutter issues~\cite{Liu2017}. ParetoLens capitalizes on the visual analytics paradigm, enabling human analysts to interact effectively with a large number of solutions and reference points. This facilitates the disclosure of patterns hidden in the distribution of solutions and aids in identifying potential performance issues.

\subsection{Visualization in Explainable AI}
\label{sec:related-work-xai}

A considerable emphasis in current AI research is placed on facilitating the understanding of output and internal mechanisms of AI models~\cite{LaRosa2023}. This emphasis has led to the development of the explainable AI (XAI) paradigm, which fulfills the need to comprehend how models generate outputs and when these models may fail. Some research has been conducted on visualizing the internal mechanisms of specific learning models. BaobabView~\cite{VanDenElzen2011BaobabView} and BOOSTVis~\cite{Liu2017b} employ tree visualization techniques to illustrate how decisions are made in the models. 
Similarly, in deep neural networks, several frameworks offer interactive environments for understanding data flow across network layers and tracking model evolution throughout training epochs, such as CNN Explainer~\cite{Wang2021}, DeepTracker~\cite{Liu2018a}, and GAN Lab~\cite{Kahng}.

Parallel to the development of visualizing model internals, there has been considerable work on providing a model-agnostic method of presenting output data from models~\cite{yuan2021survey,LaRosa2023}. Such approaches enhance the generalizability of visualization methods as they unify different model implementations under the same data abstraction protocol. For example, the Manifold framework proposed by Zhang \textit{et al.}~\cite{Zhang2018} supports the analysis of predicted labels from classification models, enabling visual model comparison and feature attribution irrespective of the model types. RuleMatrix~\cite{Ming2018} extracts classification rules based on predictions of a given unlabeled dataset to reveal the main decision criteria of the model, while Yedjour et al.~\cite{YEDJOUR2024111160} utilized multi-objective optimization approaches to extract explainable rules that both satisfy fidelity and accuracy requirements. The work by Alsallakh \textit{et al.}~\cite{Alsallakh2014} provides a visual abstraction of probabilistic classifiers to depict predictions and their corresponding uncertainties.

Given the successful application of interactive visualization in XAI, it is posited that using visualization to explain and diagnose complex models or algorithms can significantly enhance the inspection and exploration of solution sets. Our framework adopts a model-agnostic design, focusing on solution sets regardless of the specific MOEAs used to generate it. This approach allows our framework to be generalized across a wide range of MOEAs.

\section{Overview}

In this section, the research principles underlying the development of the visual analytics framework are presented in the context of this study. Subsequently, a user-centric design process is outlined, emphasizing the incorporation of domain experts in evolutionary computing. Finally, a compilation of analytical tasks is distilled to direct the design of the framework's functionalities.

\begin{figure}[t!]
    \centering
	\includegraphics[width=\columnwidth]{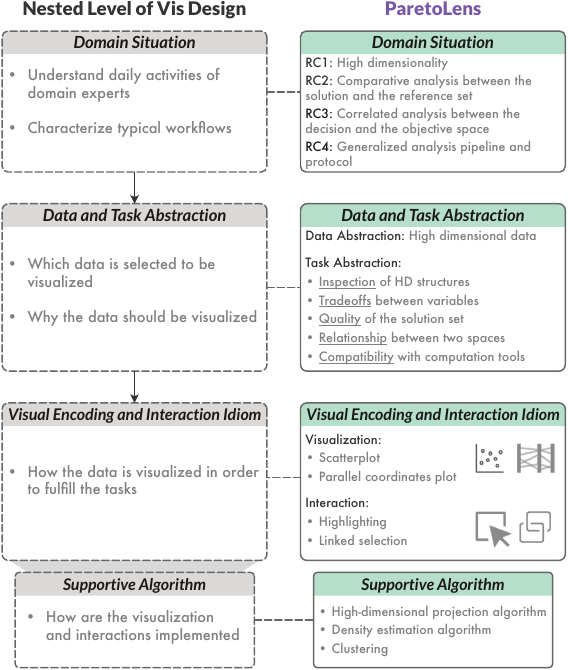}
	\caption{The nested model for visualization design~\cite{Munzner2009,Munzner2014} is shown on the left side. The corresponding design stages regarding this work are illustrated on the right side.}
	\label{fig:nested_model}
\end{figure}

\subsection{Research Guideline}

With respect to the design of our visual analytics framework, the well-established nested model in visualization research~\cite{Munzner2009,Munzner2014} is followed, serving as a guideline for developing the framework. As illustrated in Figure~\ref{fig:nested_model} (Left), the nested model consists of four levels:

\begin{enumerate}[leftmargin=*]
    
    \item \textbf{Domain Situation.} This level is to comprehend the working routines of domain experts in their daily activities and the characteristics of typical workflows. It is worth noting that the term \textit{domain} is employed explicitly in our paper to denote the field encompassing the target users of our framework, i.e., MOEA.
    
    \item \textbf{Data and Task Abstraction.} Within this level, the major goal is to identify the domain situation that ``why'' visualization is necessary, i.e., the analytical tasks that users may execute. Furthermore, considerations about which data require visualization are also considered.

    \item \textbf{Visual Encoding and Interaction Idiom.} 
    Drawing from the data and tasks determined in the previous level, this level addresses ``how'' such analysis can be fulfilled. A design for both the visualizations and the corresponding interactions is to be provided.

    \item \textbf{Supportive Algorithm.} This level is closely connected to the previous one, specifying the details of how the designed visualizations and interactions are computationally implemented.
\end{enumerate}

In the context of our research in this work, the core concepts regarding the four levels are mapped to their respective components in MOEAs, Figure~\ref{fig:nested_model} (Right). The corresponding details of the levels are described in the following sections.

\begin{figure*}[t!]
    \centering
	\includegraphics[width=\textwidth]{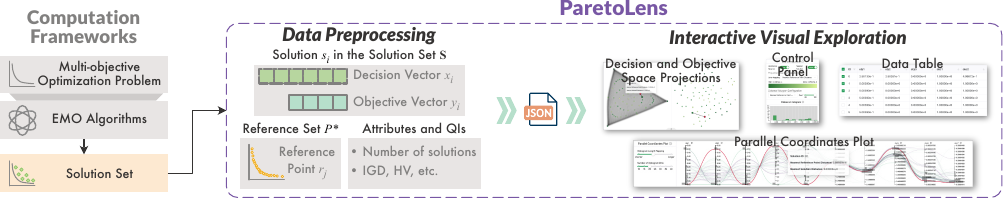}
	\caption{The overview of the framework with two main stages. After the required data about solutions and reference sets are exported from the computation framework, the \textit{data preprocessing stage} converts the collected data into an intermediate transmission format in JSON. The file is then loaded into the visualization frontend to enable the \textit{interactive visual inspection and exploration} of solutions.}
	\label{fig:overview}
	\vspace{-3mm}
\end{figure*}

\subsection{Preliminary Study of Domain Experts}
\label{sec:preliminary_study}

During the design stage of our framework, we closely collaborated with two domain experts in the evolutionary computing community in order to better understand the requirements. The first expert (denoted as Expert 1, also one of the co-authors) possesses nearly 15 years of professional research experience in evolutionary computing and multi-objective optimization. The second expert (denoted as Expert 2) has over ten years of experience in multi-objective optimization and machine learning. The domain experts frequently incorporate MOEAs as a part of their routine research workflows. In the two-stage design process, the first stage consists of multiple semi-structured interviews to gather insight into the domain-specific background of MOEA and analytical tasks closely related to the experts' workflow. After obtaining the initial insights, the framework was then designed and refined iteratively over the range of 6 months through bi-weekly to monthly engagements with the experts until functionality and user experience issues were fully tackled.

During the collaboration with the domain experts, several \textbf{research challenges (RC1-RC4)} were identified, and varying analytical tasks that should be addressed in the visual analytics framework were distilled. Such insight also aligns with the first two levels in the design guideline mentioned above.

\begin{itemize}[leftmargin=*]
\item \textbf{RC1: High dimensionality.} The number of dimensions to both the decision and the objective spaces frequently exceeds the human capacity for visual perception, typically constrained to a two- or three-dimensional visible space. Such limitation poses challenges in the straightforward visualization of solutions and the effective recognition of salient patterns hidden in the data distributions~\cite{Liu2017}. Consequently, addressing the effective delivery of high-dimensional structures is essential in the design of our visual analytics framework.

\item \textbf{RC2: Comparative analysis between the solution and the reference set.} Based on the taxonomy of evaluation aspects in Li \textit{et al.}~\cite{Li2019}, effective presentation of the relationships between solutions and reference points provided in many benchmarking problems facilitates the inspection of solution qualities from multiple aspects. For example, the experts emphasized that visualizations should be helpful in identifying solution spread and convergence towards the Pareto front by utilizing the reference set for context.

\item \textbf{RC3: Correlated analysis between the decision and the objective space.} Existing evaluation methods mainly focus on analyzing solutions in the objective space. However, the experts noted that understanding relationships between solutions and their decision vectors can unveil test problem characteristics. For instance, a cluster in the objective space might display a scattered pattern across the decision space, and solutions with increasing objective values may not show a continuous trend in the decision space.

\item \textbf{RC4: Generalized analysis pipeline and protocol.} Instead of examining the internal mechanisms of individual algorithms, the experts expressed interest in constructing a transparent, universally applicable analysis pipeline for the majority of MOEAs. Additionally, several frameworks and libraries (e.g., pymoo~\cite{pymoo}, PlatEMO~\cite{PlatEMO}) have implemented numerous algorithms across diverse programming languages and platforms. The proposed visual analytics framework should accommodate outputs from various computational tools, thereby supporting broader applicability.

\end{itemize}

\subsection{Data and Task Analysis}
\label{sec:task_analysis}

Following the identification of research challenges essential for comprehending the domain situation, the abstraction of data and tasks involved in the framework design is derived.

\subsubsection{Data Abstraction}
\label{sec:data_abstraction}
The data types for the proposed visualization framework are derived from the definitions associated with multi-objective optimization settings. Multi-objective optimization problems can be conceptualized as searching for extreme values for $m$ objective functions (which span an $m$-dimensional objective space, $Y$) in a decision space $X \in \mathbb{R}^n$ comprising $n$ decision variables. The solutions computed from a single algorithm run are represented as $s_i = (\mathbf{x}_i, \mathbf{y}_i)$, with the decision vector $\mathbf{x}_i$ belonging to $X$ and the objective vector $\mathbf{y}_i$ belonging to $Y$. The union of all the solutions $s_i$ forms a solution set $\mathbb{S}$. Our framework adopts an abstraction where the data to be visualized is considered high-dimensional, including both the decision and the objective vectors from a solution set. Consequently, high-dimensional visualization techniques are well-suited for such data. Furthermore, for certain benchmarking problems, a reference set $\mathbf{P^*} = \{r_j \in Y\}$ is often provided, which serves as a nearly-accurate representation of the Pareto front. In practice, such reference is extensively used in the quality assessment of solutions from algorithmic runs.

\subsubsection{Analytical Tasks}
Based on the four research challenges outlined in Section~\ref{sec:preliminary_study}, the subsequent \textbf{analytical tasks (T1-T4)} were further distilled to inform and guide the design of visualizations and interactions in the framework.

\begin{itemize}[leftmargin=*]
    \item \textbf{T1: Facilitate visual inspection of high-dimensional structures in the solution set.} In accordance with RC1, the analysis of data distributions and insightful patterns constitutes the primary target when evaluating solutions from individual algorithm runs.
    
    \item \textbf{T2: Depict trade-offs and correlations between solution values on decision and objective variables.} Following RC1, in addition to examining distribution information, it is essential to demonstrate how objective values of solutions shift in response to changes in other variables, including trade-offs between objectives and correlations among objective and decision variables.

    \item \textbf{T3: Represent the quality of the solution set.} To assess solution quality, as indicated by RC2, visual comparisons should be employed to elucidate the positioning of solutions in relation to the reference set. Multiple aspects of quality assessment should be considered.
    
    \item \textbf{T4: Uncover relationships between decision and objective vectors of solutions.} Interactive and coordinated visualization techniques should be utilized to facilitate the understanding of connections between the distribution of decision vectors and their corresponding objective vectors, as per RC3.
    
    \item \textbf{T5: Ensure compatibility with widely-used computation frameworks.} In line with RC4, our implementation should be designed to be compatible with existing MOEA computing frameworks.
\end{itemize}

\section{Visual Analytics Framework Design}
\label{sec:framework}

The data and task abstraction findings informed the framework design, as depicted in Figure~\ref{fig:overview}. The framework contains two main stages:

\vspace{1.2mm}\noindent\textbf{Data Preprocessing.} In this stage, a computation framework is initially deployed to generate the output of a chosen MOEA for a test problem. The output of the algorithm run is then processed to conform to the input requirements of the visualization components.

\vspace{1.2mm}\noindent\textbf{Interactive Visual Exploration.} To facilitate the exploratory data analysis of solutions, the following visualizations are incorporated:

\begin{itemize}[leftmargin=*]
    \item The \textbf{decision space projection} enables analysts to assess the distribution of decision vectors for solutions produced during an algorithm run. Moreover, the \textbf{objective space projection} visualizes the corresponding objective vectors. Additional interactive filtering and highlighting techniques are utilized to reveal solution characteristics in both projections, which is controlled in the \textbf{control panel} on the left side.
    
    \item For examining solution values across various decision and objective variables, the \textbf{parallel coordinates plot (PCP)} and the \textbf{data table} are designed to support the inspection of trade-offs and correlations.
\end{itemize}

\begin{figure*}[t!]
    \centering
    \includegraphics[width=\textwidth]{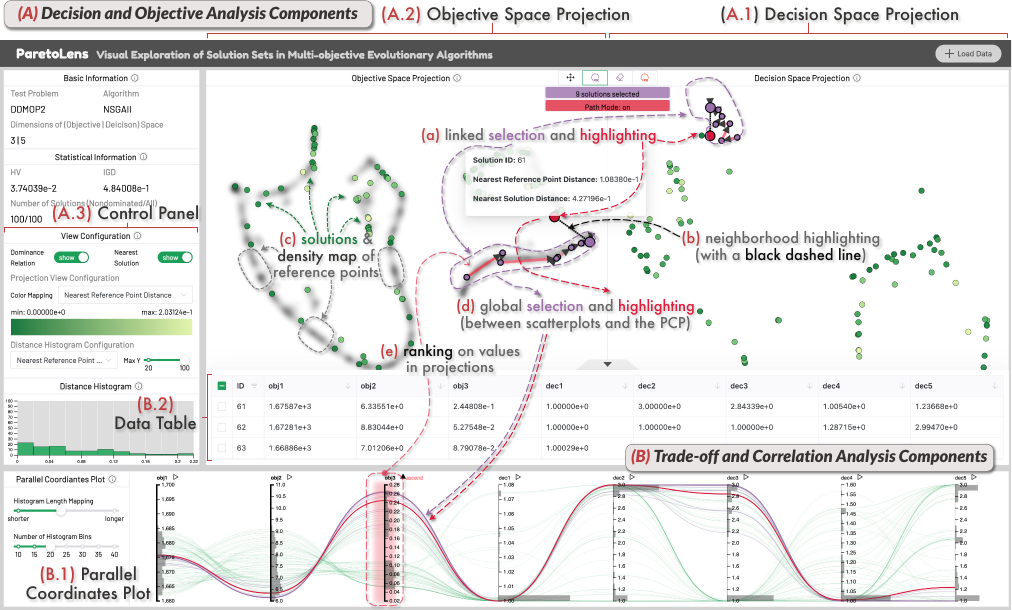}
    \caption{The interface of ParetoLens comprises a set of visualization components (A.1-A.3, B.1-B.2) aimed at visually inspecting and exploring solutions from two major analytical aspects: (A) decision and objective space analysis, and (B) trade-off and correlation analysis.}
    \label{fig:interface_teaser}
    \vspace{-2mm}
\end{figure*}

Figure~\ref{fig:interface_teaser} presents the interface of the proposed visual analytics framework. Analysts can transition between different visualization components in their analysis workflow. The interface is implemented in a browser-server architecture, allowing analysts to utilize the framework via a web browser. Further details regarding the implementation of the framework can be found in Section~\ref{sec:system_implementation}.

\subsection{Data Preprocessing}
\label{sec:data_preprocessing}
As highlighted in \textbf{RC4}, our framework is designed to accommodate a variety of MOEAs with different internal evolutionary strategies. Thus, the initial stage of the framework is designed for transforming outputs from computational frameworks and libraries into a common intermediate representation. This representation encapsulates key information of multi-objective optimization as summarized in the data abstraction section (Section~\ref{sec:data_abstraction}), namely, the corresponding decision and objective vectors of all solutions. The widely-used JSON data format is used to implement the representation for exchanging data between the backend server and the visualization front-end in the web browser environment. It should be noted that the framework also records associated metadata of the test problem, such as the count of solutions and dimensions of the decision and the objective spaces, which could facilitate rapid recall and reference in the exploration process. Some affiliated data utilized in the interactions and visualizations are included in the transferred data as well.

\subsection{Decision and Objective Space Analysis}
In order to depict the distribution of the solutions in the decision and the objective spaces, our framework employs a projection-based visualization design with a rich set of visual encodings and interactions. The visualizations are intended for fulfilling the analytical tasks of inspecting high-dimensional structures (\textbf{T1}), quality assessment (\textbf{T3}), and correspondence between decision and objective vectors (\textbf{T4}). Illustrated in Figure~\ref{fig:interface_teaser}, three major components are associated with this analysis stage: two projection components (A.1 and A.2 in Figure~\ref{fig:interface_teaser}) for the decision and the objective vectors, respectively, and an affiliated control panel (A.3 in Figure~\ref{fig:interface_teaser}) for adjusting the visual features in the projections.

\vspace{1.2mm}\noindent\textbf{Basic Design.}
The solutions of an algorithmic run are visualized in scatterplots, where each point denotes an individual solution. The coordinates of these points are determined by applying the t-SNE~\cite{vanDerMaaten2008} or the UMAP~\cite{mcinnes2018umap} algorithm to both the decision and the objective vectors of all the solutions respectively, thereby yielding two separate panels of projections.\footnote{For details of the algorithms, please refer to the supplementary material.} It is worth noting that the objective vectors corresponding to the reference set are appended when generating the projections of the objective space, which ensures alignment between the coordinates of the the solutions and the reference points.

\vspace{1.2mm}\noindent\textbf{Enhanced Visual Encodings and Interactions.}
Depicted in (A.1) and (A.2) of Figure~\ref{fig:interface_teaser}, a straightforward method of presenting projection results is through a conventional static scatterplot. However, with an increase in the number of solutions and reference points, the visualization tends to become heavily cluttered, rendering it impossible to read, as highlighted by recent studies in the visualization community~\cite{Sarikaya2018,Yuan2021a}. To enable an in-depth examination of the results, various enhancements have been incorporated to facilitate visual exploration of the projection results.

\begin{itemize}[leftmargin=*]
	\item \textit{Linked Selection}: Two interactive selection mechanisms are employed to emphasize selected solutions: hovering and lasso selection. Positioning the mouse pointer over a specific point in either scatterplot enables a linked highlighting of the corresponding solution in the other scatterplot, Figure~\ref{fig:interface_teaser} (a), where both highlighted points are rendered red. At the same time, a tooltip is activated next to the point, which presents detailed information regarding the solution. The information shown in the tooltip includes a unique identifier for the solution as well as the distances to the closest reference point and solution in the objective space.

    \begin{figure}[t!]
        \centering
        \includegraphics[width=\columnwidth]{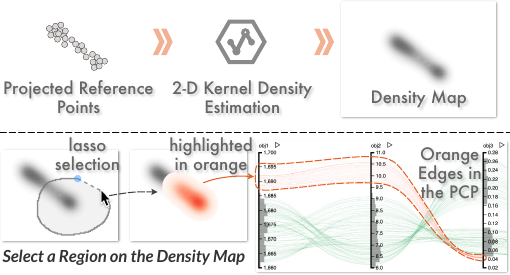}
        \caption{(Top) Design of the density map of reference points. The density map is generated with 2-D kernel density estimation applied on all reference points. (Bottom) Lasso-selection on the density map. The selected reference points render a new density map patch in orange; they are also displayed in the PCP as orange edges among the axes for objectives.}
        \label{fig:kde_illustration}
        \vspace{-4mm}
    \end{figure}

	\item \textit{Density Map of Reference Points}: In many commonly utilized benchmarking problems~\cite{Deb2002DTLZ,He2020}, the number of reference points can be substantial, leading to significant visual clutter in the conventional scatterplot setting. To alleviate the issue of over-plotting, the scatterplots are optimized using the 2-D Kernel Density Estimation (KDE) method to approximate the density of the reference set distribution in the projection result as shown in Figure~\ref{fig:kde_illustration} (Top). Specifically, the Gaussian kernel function is adopted during the density estimation process. The resulting density is then rendered as a gray-scale density map under the solutions, Figure~\ref{fig:interface_teaser} (c), where a darker shade indicates a higher density. In this way, the density map offers an appropriate context for conducting comparative analysis between the solutions and the established reference points in the test problem. Furthermore, when an area is lasso-selected on the density map, the corresponding region is highlighted. In Figure~\ref{fig:kde_illustration} (Bottom), a new density map patch, computed using reference points inside the selected region and displayed in orange, is appended to the default density map. The corresponding reference points are also presented as orange edges in the PCP. To ensure that outlying reference points are not overlooked, an outlier preservation step is applied after rendering the density maps. This step involves the identification of potential outliers in the reference set using the Local Outlier Factor method~\cite{lof2000}. The outliers are then explicitly depicted in the objective scatterplot as smaller gray dots, thereby ensuring they remain visible while being distinguishable from the main data set.

	\item \textit{Neighborhood Highlighting}: A recognized challenge associated with non-linear projection techniques is the distortion of distance relationships among data instances in the projected 2-D visualization outcomes, potentially leading to misidentification of nearest neighbors~\cite{Liu2017}. In the framework, an explicit visual encoding scheme is applied to denote an individual solution's nearest neighbor by linking a black dashed line between them, Figure~\ref{fig:interface_teaser} (b). This link becomes visible when a solution in the objective scatterplot is hovered upon with the mouse pointer. A toggler in the control panel is designed to enable or disable this feature.

\begin{figure}[t!]
    \centering
	\includegraphics[width=\columnwidth]{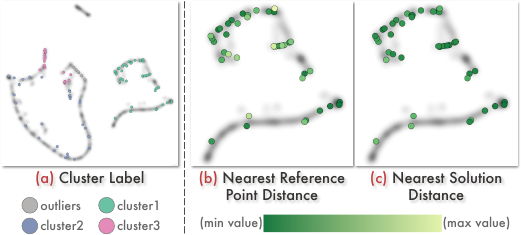}
	\caption{Color mapping of the scatters in the projections.}
	\label{fig:projection_color_mapping}
	\vspace{-3mm}
\end{figure}

	\item \textit{Clustering}: Following the aforementioned challenge, intrinsic cluster structures hidden in the objective space may be obscured in the projection results as well. To address this issue, the HDBSCAN algorithm~\cite{mcinnes2017accelerated} is first applied to the objective vectors of the solutions. HDBSCAN, a density-based hierarchical clustering algorithm, computes mutual density distances to construct a minimum spanning tree, connecting data points with similar densities. This process results in a hierarchical cluster tree, from which the most stable clusters are selected based on persistence. The algorithm automatically determines the optimal number of clusters and efficiently handles noise.
    Then, the cluster labels are mapped to the colors of the points in the objective scatterplot, Figure~\ref{fig:projection_color_mapping} (a), where a categorical color scheme is employed. Outliers identified by the algorithm are represented in a gray hue. Users have the option to control the display of cluster labels in the ``View Configuration'' panel.

	\item \textit{Color Mapping of Measures}: In addition to using point colors to present the clusters, alternative color mapping strategies have been designed to illustrate the attributes of the solutions. In the control panel in Figure~\ref{fig:interface_teaser} (A.3), two additional options under the ``Color Mapping'' selection are provided to facilitate the continuous color mapping of ``Nearest Reference Point Distance'' and ``Nearest Solution Distance'', which is illustrated in Figures~\ref{fig:projection_color_mapping} (b) and~\ref{fig:projection_color_mapping} (c). A color bar is provided where smaller distance values are assigned to lower lightness of the green color. These two options can expose the quality of solutions on a per-solution basis. Moreover, a histogram of the distances in the control panel, ``Distance Histogram'' in Figure~\ref{fig:interface_teaser} (A.3), allows solution filtering based on distance values. By brushing along the horizontal axis, solutions corresponding to the selected distance value range will be emphasized in the scatterplots with dark strokes.
	\item \textit{Dominance Relationship}: In certain MOEAs that do not employ post-hoc non-dominated sorting actions, it is possible for dominated solutions to persist in the resulting solution set. To address this, an additional non-dominated sorting step is applied to the input solution set to identify the dominated solutions. A toggler called ``Dominance Relation'', Figure~\ref{fig:interface_teaser} (A.3), is incorporated into the control panel. This feature allows for altering the visual representation of dominated solutions in the projections by changing their shape from circular dots to crosses, which emphasizes the presence of dominated solutions.
\end{itemize}

\subsection{Trade-off and Correlation Analysis}

Despite the distribution patterns shown in the projections, a critical limitation in the projection results is their inability to depict the precise values of the solutions. As mentioned in \textbf{T2} and \textbf{T4}, a supplementary perspective is offered in the PCP, which portrays exact values through axis-based visual encoding. Additionally, the data table, Figure~\ref{fig:interface_teaser} (B.2), enumerates all values in a tabular representation.

Shown in Figure~\ref{fig:interface_teaser} (B.1), the PCP contains an enhanced PCP and an affiliated control panel.

\vspace{1.2mm}\noindent\textbf{Basic Design.} Given the high-dimensional characteristics previously discussed, PCP is adopted, which displays each variable with a vertical axis and each solution with an edge. The axes for both spaces are incorporated into the same plot to reveal the relationships between decision variables and objectives, as outlined in \textbf{T4}.

\vspace{1.2mm}\noindent\textbf{Enhanced Visual Encodings and Interactions.} While PCPs are extensively employed in visualizing high-dimensional data, the typical design remains plagued by various issues, including visual clutter and over-plotting~\cite{Sarikaya2018}. Furthermore, using linked analysis across multiple visualization components can enhance the multi-faceted examination of the solutions and reference points. As such, several improvements to the PCP are applied.

\begin{itemize}[leftmargin=*]
	\item \textit{Brushing and Hovering}: This set of interactions are applied to mitigate the visual clutter caused by massive edges, i.e., solutions. Brushing on one or more axes emphasizes the edges that match the selected value intervals; hovering over an edge facilitates a more detailed emphasis of an individual solution.
	
	\item \textit{Density Histogram on Axes}: Density histograms are attached to the right side of all axes to reveal the density of edges passing through a given axis along its own value range. The quantity and the maximum visual heights of bins can be adjusted in the control panel on the left side.
	
	\item \textit{Axes Reordering}: To explicitly illustrate the correlations between two variables, as indicated in \textbf{T2}, axes can be dragged and reordered, facilitating pairing any two axes. In this way, the distribution patterns of edges between two adjacent axes may expose positive, negative, or potentially more complicated correlations of the corresponding variables.
\end{itemize}

\begin{figure*}[t!]
    \centering
	\includegraphics[width=\textwidth]{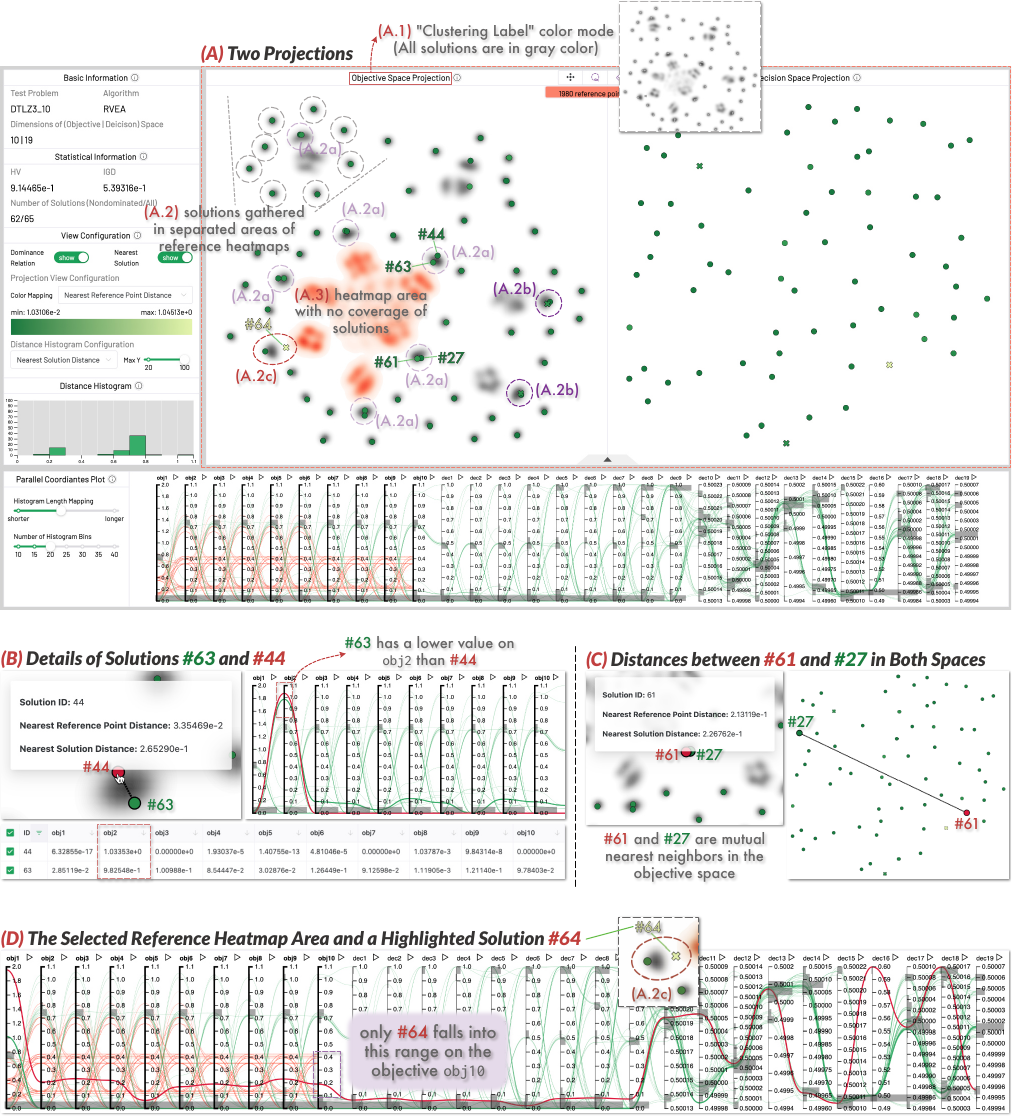}
	\caption{The result of RVEA on the DTLZ3 problem in Case Study 1. (A) A snapshot of the main interface. In the objective space projection result, Alice has made several noteworthy observations regarding distributions, color mappings, and the reference density map. (B) A detailed description of the dominance relationship between two solutions (\#63 and \#44). They are considered mutual nearest neighbors and cannot dominate each other. (C) A detailed illustration of a pair of mutual nearest neighbors. They are closely positioned in the objective space while maintaining a considerable distance from each other in the decision space. (D) The PCP when a reference density map area shown in (A.3) is selected, with solution \#64 highlighted.}
	\label{fig:case1}
	\vspace{-3mm}
\end{figure*}

\subsection{Cross-component Coordinated Interactions}

Besides the interactions described above, our framework supports interactions that are responded to by multiple visualization components in the framework. 

\begin{itemize}[leftmargin=*]
	\item \textit{Global Selection and Highlighting}: The selection and highlighting interactions in the two projections and the PCP are intrinsically interconnected, as Figure~\ref{fig:interface_teaser} (d) demonstrates. Activating solutions in one visualization component will illuminate the corresponding visual elements in the other components. A particularly representative scenario arises when selecting edges by brushing on one or multiple axes. The corresponding point distributions in the projections can be insightful regarding the global similarity of the solutions.

	\item \textit{Ranking on Values in Projections}: A limitation inherent to the projection results is the distortion of axis information, a necessary trade-off to preserve similarity relationships. To simultaneously show how solutions are ranked based on their variable values while also portraying the similarities, a sequence of connected arrows can be toggled in the projection results. These arrows indicate the increase in value across the selected solutions. As illustrated in Figure~\ref{fig:interface_teaser} (e), clicking on the small triangular icon next to the axis name alternates between different display modes of arrows in the projection results, including pointing from the solution with the smallest value to the one with the largest value on the variable as well as turning off this feature.
\end{itemize}

\section{Evaluation}

This section describes how ParetoLens enhances the inspection and exploration of solution sets through three representative case studies. The first case study highlights the tool's fundamental functionalities, while the subsequent two focus on its applicability to real-world problems. The data preprocessing performance of ParetoLens is further assessed through a quantitative evaluation across different data scales. Additionally, an expert interview was conducted to gather feedback from the target users of the visual analytics framework. 

To better illustrate the analytical processes in the case studies, a fictional analyst named Alice is introduced, who is tasked with analyzing solution sets derived from various established MOEAs.

\subsection{System Implementation}
\label{sec:system_implementation}
The framework is implemented by following the browser-server architecture, facilitating effortless access via common web browsers. As mentioned in Section~\ref{sec:data_preprocessing}, PlatEMO~\cite{PlatEMO} is employed to execute algorithm runs and generate the solution sets. The web server is powered by Python Flask\footnote{\url{https://flask.palletsprojects.com}}, while the frontend UI and visualizations are built with Vue3\footnote{\url{https://vuejs.org}} and d3.js~\cite{bostock2011d3}, respectively. 
The following hyperparameter settings are currently employed in the algorithms: \texttt{perplexity}=30 and \texttt{learning\_rate}=200 in t-SNE, \texttt{n\_neighbors}=15 and \texttt{min\_dist}=0.1 in UMAP, and \texttt{min\_cluster\_size}=10 and \texttt{min\_samples}=2 in HDBSCAN.

\subsection{Case Study 1: DTLZ3 Test Problem}

In the first case study, the DTLZ3 test problem is employed as an illustrative example to navigate through the fundamental functionalities provided in the interface. This problem is characterized by 19 decision variables and 10 objectives. The RVEA algorithm~\cite{Cheng2016RVEA} is selected to demonstrate our framework, with UMAP employed for projections in both the decision and the objective spaces.

After importing the data into the interface, Alice begins her analysis of the solution set by examining the data distributions in the projections. The objective space projection reveals a relatively uniform distribution of solutions without noticeable clusters, Figure~\ref{fig:case1} (A). This observation is further confirmed when the solution color mapping is set to ``Clustering Label,'' which reveals a uniform gray color, indicating the absence of significant clusters, Figure~\ref{fig:case1} (A.1).

\begin{figure}[t!]
    \centering
	\includegraphics[width=\columnwidth]{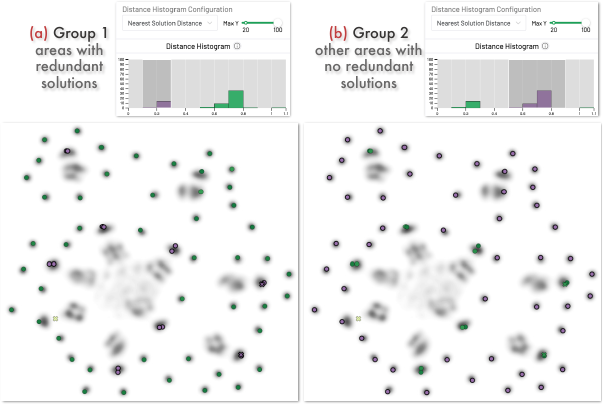}
	\caption{The corresponding selections of solutions in the distance histogram. (a) Group 1: The selected solutions are associated with areas where duplicated solutions exist within each reference density map area. (b) Group 2: The majority of solutions with average distances are included in this group.}
	\label{fig:case1_histogram}
\end{figure}

\begin{figure*}[t!]
    \centering
	\includegraphics[width=\textwidth]{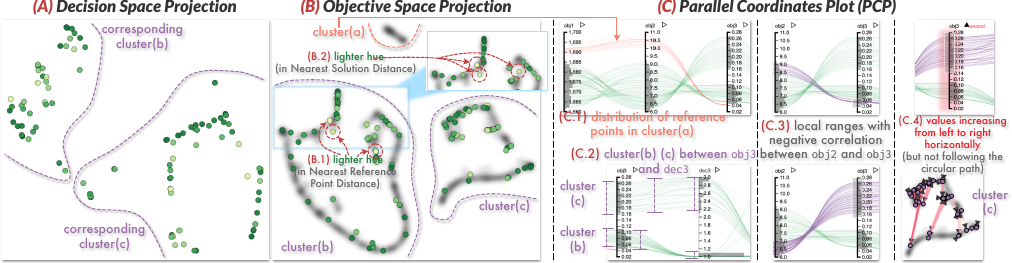}
	\caption{The result of NSGA-II on the DDMOP2 problem in Case Study 2. (A+B) The decision and the objective space projections are depicted on the left side and in the middle of the figure, respectively. In the objective space projection result, three clusters, namely (a), (b), and (c), can be observed, while cluster (a) lacks solution coverage. Some solutions also show low values with respect to nearest reference point distances or nearest solution distances (C) The snapshots of the PCP are listed.}
	\label{fig:case2}
	\vspace{-3mm}
\end{figure*}

Further exploration by Alice uncovers that most solutions are gathered in a separated density map area of reference points, although certain areas contain multiple solutions within the same area. In certain areas labeled (A.2a) where more than one solution falls into, solutions are mutually non-dominant; meanwhile, in those areas marked (A.2b), solutions dominated by their counterparts can be identified. Specifically, \#64, the dominated one in area (A.2c), shows considerably low quality with respect to the distance to the nearest reference point. Taking the relationship between solutions \#44 and \#63 as an example, Alice notices that these solutions are mutual nearest neighbors in the objective space, which is indicated by the dashed lines when hovering on the two solutions respectively. Depicted in Figure~\ref{fig:case1} (B), a detailed comparison in the PCP shows that solution \#63 depicts a lower value on the objective \texttt{obj2}, whereas \#44 outperforms in all other objectives with lower values. This observation can also be reflected in the data table by comparing their objective values.

Activating the Distance Histogram in ``Nearest Solution Distance'' mode, Alice identifies two distinct groups of bars in the histogram. As illustrated in Figure~\ref{fig:case1_histogram}, the bars in Group 1 (Left) represent those areas with redundant solutions in close proximity, while Group 2 (Right) reflects the average distances among most solutions.

Alice observes another interesting phenomenon regarding the visual nearest-neighbor distances among solutions, represented by dashed lines. Dot pairs that seem to be close to each other in the objective space projection often show significant separation in the decision space, as shown in the example of solutions \#61 and \#27 in Figure~\ref{fig:case1} (C). This observation suggests a complex landscape in the decision space, where variations in values across different spaces do not align proportionally. This disparity between the objective and the decision spaces underscores the intricate nature of the optimization problem, highlighting the challenge of navigating the decision space to identify optimal solutions.

Furthermore, Alice notes the absence of solutions in certain reference density map areas, Figure~\ref{fig:case1} (A.3), suggesting potential gaps in coverage on these areas of the Pareto front. By using the lasso tool to select these specific reference density map areas and then focusing on the axis \texttt{obj10} through brushing, Alice observes a notable gap in solution coverage (particularly in the range between 0.1 and 0.4), except for an outlying dominated solution (solution \#64), further highlighting the issue of inadequate front coverage.

\subsection{Case Study 2: DDMOP2 Test Problem}
In the second case study, the DDMOP2 test problem from the DDMOP test suite~\cite{He2020} is employed, where the problems are devised from scenarios encountered in real-world applications. This problem comprises five decision variables, symbolizing the reinforcing components of automobile frontal structures while aiming to minimize three objectives.

Upon loading the requisite data for the runs of existing algorithms, Alice contemplates the efficacy of standard MOEAs in addressing this test problem. Consequently, she opts to examine the results obtained from NSGA-II, as depicted in Figure~\ref{fig:case2}. The objective space projection in the middle reveals approximately three clusters based on the density map of reference points. However, it is noted that one cluster, labeled (a) in orange color, lacks solutions (i.e., solid dots), suggesting that the algorithm might not have successfully identified the corresponding region of the Pareto front. Through utilizing the lasso selection tool on the density map, Alice observes the absence of solutions within the ranges of orange edges for objectives \texttt{obj1} and \texttt{obj2} in the PCP, Figure~\ref{fig:case2} (C.1), further highlighting the limitations of this solution set.

Alice then proceeds to inspect the quality of the existing solutions. By applying the ``Nearest Reference Point Distance'' color mapping, she identifies some solutions marked by significantly lighter hues in cluster (b) in the objective space projection result, Figure~\ref{fig:case2} (B.1), indicating their relative distance from the reference points and potentially low quality. Switching the color mapping to ``Nearest Solution Distance'' reveals several solutions with notably greater distances, as shown in the light blue rectangle in Figure~\ref{fig:case2} (B.2). This observation may serve as a valuable aid in 2-D projections where such distances are less apparent. Moreover, these solutions may suggest a considerable sparsity in the corresponding regions of the objective space. To inspect deeper into the relationships between the objective and the decision vectors of the solutions, Alice employs the lasso selection tool to highlight the two previously identified clusters, i.e., clusters (b) and (c). In Figure~\ref{fig:case2} (C.2), the distribution patterns of the objective vectors are found to correspond with those of the decision vectors in terms of clustering patterns and relative positioning along a specific path, suggesting potential correlations between decision and objective variables.

\begin{figure}[t!]
    \centering
	\includegraphics[width=\columnwidth]{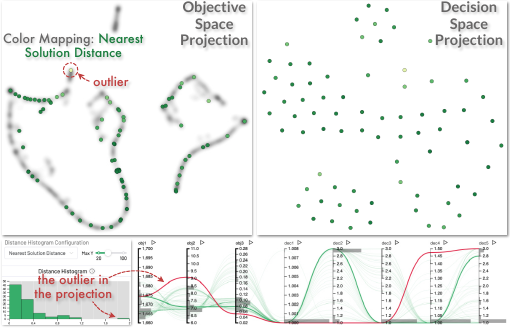}
	\caption{The result of MOEA/D on the DDMOP2 problem in Case Study 2. Compared with the results from NSGA-II in Figure~\ref{fig:case2}, MOEA/D provides a more uniform distribution of solutions with no obvious clustering patterns. In addition, an outlying solution can be discovered, marked as ``outlier'' in this figure.}
	\label{fig:case2_comparison}
	\vspace{-3mm}
\end{figure}

Upon reviewing the projection results, Alice aims to investigate the value distributions across decision and objective variables. The PCP result in Figure~\ref{fig:case2} (C.2) reveals that for \texttt{obj3}, the edges span two distinct ranges. Selecting these ranges individually, Alice identifies that \texttt{obj3} plays a central role in cluster formation. This clustering pattern is similarly observed with the decision variable \texttt{dec3}. By adjusting the position of the \texttt{dec3} axis adjacent to \texttt{obj3}, a roughly positive correlation is noted in the cluster with larger values, whereas solutions within the other cluster exhibit closely aligned values on \texttt{dec3}. Additionally, near-negative correlation patterns are identified between \texttt{obj3} and \texttt{obj2} over localized ranges, shown in Figure~\ref{fig:case2} (C.3). Moreover, when Alice activates the value ranks in projections for objectives, such as \texttt{obj3} in Figure~\ref{fig:case2} (C.4), a trend of red arrows traversing from one side to the other within both clusters is observed, indicating that solution values do not increase along the clusters' path-like contour but rather from one side to the other.

Following her analysis of NSGA-II results, Alice decides to explore outcomes from alternative algorithms. Upon loading results from the MOEA/D algorithm (Figure~\ref{fig:case2_comparison}), notable differences can be discovered. First, changing the solution color mapping to ``Nearest Solution Distance'' in the projections results in a more uniform color distribution compared to that observed with NSGA-II. The histogram of distances to the nearest solutions (at the bottom left corner of Figure~\ref{fig:case2_comparison}) mainly falls in the first interval, suggesting a more uniform solution distribution. This uniformity is also reflected in the decision space projection result. Then, unlike the objective space, the decision vectors do not exhibit obvious clustering patterns, yet solutions within the same cluster remain relatively proximate, as determined by lasso-selecting the clusters in the objective space projection. Additionally, an outlier solution is identified, distinguished by its position and the light hue under the ``Nearest Solution Distance'' color mapping, signifying its outlyingness from the cluster.

\begin{figure}[t!]
    \centering
	\includegraphics[width=\columnwidth]{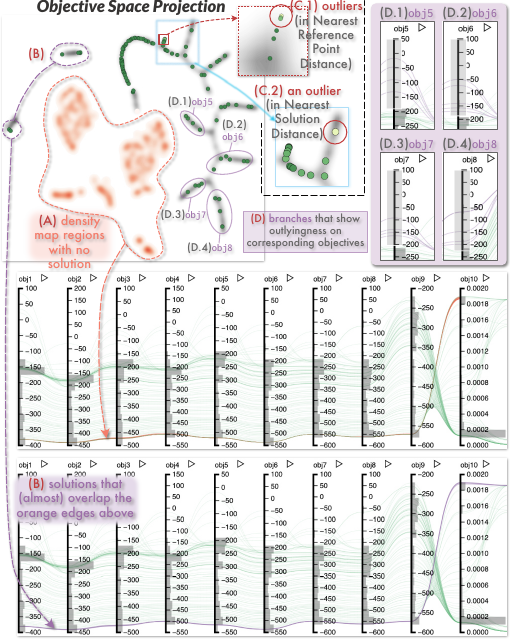}
	\caption{The results of NSGA-II on the DDMOP4 problem in Case Study 3. (A) A reference density map area lacks covered solutions; however, several surrounding solutions in (B) exhibit adequate coverage according to the PCP. (C) Several outliers can be observed. (D) The projection result illustrates several branches that depict outlying characteristics on specific corresponding objectives.}
	\label{fig:case3}
	\vspace{-3mm}
\end{figure}

\subsection{Case Study 3: DDMOP4 Test Problem}

In the third case study, the DDMOP4 test problem is used to demonstrate the capabilities of ParetoLens in addressing challenges posed by a relatively larger number of decision variables and objectives, which exceed the confines of visible spaces (i.e., three-dimensional). This problem encompasses 13 decision variables and 10 objectives based on the design of LTLCL switching ripple suppressors with nine resonant branches~\cite{He2020}.

Alice opts to analyze the results obtained through the application of the NSGA-II, which is referenced as a baseline in the literature~\cite{He2020}. As illustrated in Figure~\ref{fig:case3} (A), a notable observation is the presence of several density map regions without covering solutions. This observation may suggest that the algorithm has not adequately identified certain areas on the Pareto fronts. To investigate the value distribution of the reference points within these regions, Alice employs the lasso tool on the density map, revealing some orange edges in the PCP. Furthermore, upon examining nearby solutions around the aforementioned regions, Alice notes that some solutions nearly overlap the orange edges, Figure~\ref{fig:case3} (B), implying that the absence of solutions in certain density map regions may result from potential distortions caused by the projection algorithm, rather than an actual lack of solutions covering those areas. When assessing the ``Nearest Reference Point Distance'' and ``Nearest Solution Distance'' through color mappings in the objective space projection, certain solutions are distinguished by their significantly lighter color (Figure~\ref{fig:case3} (C.1 and C.2)), denoting their low quality and high deviation.

Then, Alice explores the value distributions using the PCP. Illustrated in the middle part of Figure~\ref{fig:case3}, an analysis of \texttt{obj1} through \texttt{obj9} reveals a positive correlation among these axes. Conversely, the distribution on \texttt{obj10} is skewed, with many solutions clustered around a value of 0.0001, while solutions exceeding a value of 0.0002 exhibit a negative correlation with the other objectives. Moreover, Alice observes that for objectives ranging from \texttt{obj1} to \texttt{obj8}, each axis features several outlying solutions positioned at the upper end of the range. By brushing on the axes individually, it becomes apparent in the objective space projection that these outliers correspond to distinct branches. Four branches in the objective space projection result and their corresponding PCP highlighting results are illustrated through Figure~\ref{fig:case3} (D.1 - D.4).

\subsection{Quantitative Evaluation on Data Processing Speed}
\label{sec:quantitative_evaluation}

This section illustrates a quantitative assessment of the data preprocessing speed to evaluate the efficiency of the processing algorithms across varying data scales. Utilizing the test problems featured in the case studies, an analysis was conducted on the total processing time required to transform the output from MOEA libraries into JSON-formatted data,  Table~\ref{tab:performance}. This evaluation spans different scales of data, specifically focusing on varying numbers of solutions. In addition to quantifying the overall processing time, the analysis specifically isolates the time consumed during the projection stage, with the corresponding time cost marked in parentheses. The findings reveal a progressive increase in processing time correlated with the number of solutions. Notably, the projection phase emerges as the most time-intensive step in the data preprocessing stage.

\begin{table}[tbp]
    \renewcommand{\arraystretch}{1}
    \caption{The data preprocessing time \\ under different data scales in seconds.}
    \label{tab:performance}
    \centering
    \scriptsize
    \begin{tabular}{l|c|ccccc}
        \hline
        \multirow{2}{*}{Problem} & \multirow{2}{*}{\begin{tabular}[c]{@{}c@{}}Proj.\end{tabular}} & \multicolumn{5}{c}{Number of Solutions} \\ \cline{3-7} 
         &  & 100 & 200 & 500 & 1000 & 2000 \\ \hline
        \multirow{2}{*}{DTLZ3} & t-SNE & \begin{tabular}[c]{@{}c@{}}18.25 \\ (18.11)\end{tabular} & \begin{tabular}[c]{@{}c@{}}18.12 \\ (17.99)\end{tabular} & \begin{tabular}[c]{@{}c@{}}19.90 \\ (19.77)\end{tabular} & \begin{tabular}[c]{@{}c@{}}21.07 \\ (20.85)\end{tabular} & \begin{tabular}[c]{@{}c@{}}26.45 \\ (25.87)\end{tabular} \\ \cline{2-7} 
         & UMAP & \begin{tabular}[c]{@{}c@{}}14.33 \\ (14.19)\end{tabular} & \begin{tabular}[c]{@{}c@{}}4.25 \\ (4.13)\end{tabular} & \begin{tabular}[c]{@{}c@{}}4.55 \\ (4.419)\end{tabular} & \begin{tabular}[c]{@{}c@{}}4.74 \\ (4.47)\end{tabular} & \begin{tabular}[c]{@{}c@{}}6.65 \\ (5.97)\end{tabular} \\ \hline
        \multirow{2}{*}{DDMOP2} & t-SNE & \begin{tabular}[c]{@{}c@{}}3.47 \\ (3.46)\end{tabular} & \begin{tabular}[c]{@{}c@{}}4.36 \\ (4.32)\end{tabular} & \begin{tabular}[c]{@{}c@{}}5.29 \\ (5.22)\end{tabular} & \begin{tabular}[c]{@{}c@{}}7.02 \\ (6.86)\end{tabular} & \begin{tabular}[c]{@{}c@{}}12.25 \\ (11.68)\end{tabular} \\ \cline{2-7} 
         & UMAP & \begin{tabular}[c]{@{}c@{}}5.44 \\ (5.41)\end{tabular} & \begin{tabular}[c]{@{}c@{}}3.45 \\ (3.43)\end{tabular} & \begin{tabular}[c]{@{}c@{}}4.01 \\ (3.94)\end{tabular} & \begin{tabular}[c]{@{}c@{}}5.47 \\ (5.23)\end{tabular} & \begin{tabular}[c]{@{}c@{}}9.03 \\ (8.40)\end{tabular} \\ \hline
        \multirow{2}{*}{DDMOP4} & t-SNE & \begin{tabular}[c]{@{}c@{}}5.23 \\ (5.09)\end{tabular} & \begin{tabular}[c]{@{}c@{}}5.85 \\ (5.77)\end{tabular} & \begin{tabular}[c]{@{}c@{}}6.95 \\ (6.85)\end{tabular} & \begin{tabular}[c]{@{}c@{}}9.88 \\ (9.70)\end{tabular} & \begin{tabular}[c]{@{}c@{}}13.81 \\ (13.32)\end{tabular} \\ \cline{2-7} 
         & UMAP & \begin{tabular}[c]{@{}c@{}}4.58 \\ (4.52)\end{tabular} & \begin{tabular}[c]{@{}c@{}}4.69 \\ (4.62)\end{tabular} & \begin{tabular}[c]{@{}c@{}}6.07 \\ (5.97)\end{tabular} & \begin{tabular}[c]{@{}c@{}}8.08 \\ (7.88)\end{tabular} & \begin{tabular}[c]{@{}c@{}}5.67 \\ (5.17)  \end{tabular} \\
         \hline
        \end{tabular}
\end{table}

\subsection{Expert Interview}

To further assess the effectiveness of the framework, we have arranged expert interviews with six domain experts. The expert group consisted of Experts 1 and 2, who had contributed to the preliminary study identifying research challenges, along with four new experts (Experts 3, 4, 5, and 6), including two researchers and two students. Their demographic information is provided in Table~\ref{tab:experts}. The interview process started with an overview of the functionalities provided by ParetoLens, followed by a demonstration of three case studies. Then, the experts were given the opportunity to interact with the interface where pre-computed solution sets from various established MOEAs and test problems are available. During the interview sessions, free-form feedback was collected from the experts on several aspects:

\begin{itemize}[leftmargin=*]
    \item Does the proposed framework align with your current practices on analyzing solution sets? Can you identify any significant differences between employing ParetoLens for visual exploration and your conventional routines?
    \item Are the core results well-presented in the interface? What are your impressions of the visualization and interaction design?
    \item Are there any future enhancements you would suggest for the current framework?
\end{itemize}

The interviews for each expert lasted approximately from 30 to 60 minutes.

\begin{table}[tbp]
    \caption{Background of the invited new experts.}
    \label{tab:experts}
    \centering
    \begin{tabular}{ccl}
    \hline
    \begin{tabular}[c]{@{}c@{}}Expert\\ No.\end{tabular} & Role             & \multicolumn{1}{c}{Expertise}                                                                   \\ \hline
    3                                                  & Researcher       & \begin{tabular}[c]{@{}l@{}}Applying MOEAs in engineering\\ applications\end{tabular}            \\ \hline
    4                                                  & Graduate Student & \begin{tabular}[c]{@{}l@{}}Specializing in multi-objective\\ optimization research\end{tabular} \\ \hline
    5                                                  & Researcher       & EMO practitioner in circuit design                                                              \\ \hline
    6                                                  & PhD Student & \begin{tabular}[c]{@{}l@{}}Employing MOEAs in deep\\reinforcement learning algorithms\end{tabular}   \\
    \hline
    \end{tabular}
    \end{table}

\vspace{1.2mm}\noindent\textbf{Analytical Pipeline.} All experts expressed their support for using interactive visualization techniques to examine solution sets generated by MOEAs. They praised the framework's ability to depict data from various aspects, i.e., both distributions and value-centric visualization components. The experts pointed out that, unlike most systems that rely on a single, static visualization, the multi-coordinated visualization design of ParetoLens enables interactive exploration, facilitating the discovery of patterns within solution sets. Furthermore, the experts also noted the compatibility of integrating into their current research workflow. As long as the solution sets generated by optimization libraries or implementations adhere to the required input format of ParetoLens, the results can be readily imported into the framework for visual analysis. In addition to analyzing the final outcomes of algorithm runs, solution sets derived from intermediate generations can also become the analytical target in ParetoLens. Such capability enriches the analysis by providing insights into the evolutionary processes, thereby enabling the understanding of the population development over generations.

\vspace{1.2mm}\noindent\textbf{Visualization Design.} The interactive visualization design was well-received by the experts, particularly for its effectiveness in illustrating solutions and reference sets. They pointed out that the linked selection and highlighting mechanism across projection and PCPs facilitates correlating the same solutions across different visual aspects, which further enhances the analysis in a single visualization component. The experts commented on the value of comparisons between solutions and reference points in the objective space projection, which offers an intuitive understanding of areas on the Pareto front with low coverage. They also appreciated the selection feature on the density map, where comparative analysis can be conducted in the PCP by providing detailed insights into the actual values of each objective.

\vspace{1.2mm}\noindent\textbf{Suggestions on Improving the Current Design.} The experts offered several recommendations for future improvements to the framework. Inspired by the second case study, one main suggestion is the integration of functionalities that allow for explicit comparisons between multiple solution sets associated with the same test problem, enabling a more comprehensive analysis. Additionally, introducing export capabilities is proposed to facilitate the storage and sharing of analysis outcomes. Specifically, this could include exporting presented visualizations as static images and saving the sequence of analytical steps in the exploration process. Finally, the researchers who are algorithm practitioners in specific application domains raise concerns on carrying semantic meanings of the solutions in the interface.

\begin{table*}[tbp]
    \caption{A comparison between ParetoLens and existing representative works.}
    \label{tab:comparison}
    \centering
    \begin{tabular}{ccccc}
    \hline
                                Approach          & {Type}                                                 & {Visualization Design}                                                                       & \multicolumn{1}{l}{{Interactivity}} & {Coordinated Multi-view Support} \\ \hline
    \multicolumn{1}{l}{PaletteStarViz~\cite{Talukder2020star}} & algorithm                                                            & high-dimensional projection                                                                         & static                                     & no                                      \\ \hline
    \multicolumn{1}{l}{Walter \textit{et al.}~\cite{Walter2022tevc}}         & algorithm                                                            & high-dimensional projection                                                                         & static                                     & no                                      \\ \hline
    \multicolumn{1}{l}{Pymoo~\cite{pymoo}}          & \begin{tabular}[c]{@{}c@{}}computation\\ library\end{tabular}        & individual charts (scatterplots, PCPs, etc.)                                                        & static                                     & no                                      \\ \hline
    \multicolumn{1}{l}{ParetoLens}           & \begin{tabular}[c]{@{}c@{}}visual analytics\\ framework\end{tabular} & \begin{tabular}[c]{@{}c@{}}multiple coordinated visualizations\\ (projections and PCP)\end{tabular} & interactive                                & yes\\
    \hline
    
    \end{tabular}
    \vspace{-5mm}
\end{table*}

\section{Discussion}
\subsection{Design Choice between 2-D and 3-D}
In our framework, the interface predominantly utilizes 2-D visualization designs to uncover patterns in the high-dimensional decision and objective vectors. Despite the prevalence of 3-D visualizations in multi-objective optimization research literature for depicting 3-D vectors or subspace slices in high-dimensional spaces, our design choice favors 2-D representations. This decision is grounded in data visualization studies~\cite{Munzner2014,ware08book}, which suggest that adding a depth axis, in most cases, can introduce greater distortions and increase cognitive load rather than effectively enhance information conveyance. Moreover, our goal is to maintain visual consistency and adaptability across varying numbers of dimensions in decision or objective spaces. The unified design approach in the current framework supports this goal by ensuring that 2-D visualizations can accommodate diverse settings, thereby providing consistent visualization forms regardless of the dimensional complexity of the data.

\subsection{Scalability to Data and Dimensions}
As described in Section~\ref{sec:data_abstraction}, the framework design adopts a data abstraction strategy that treats the decision and the objective vectors of solutions as high-dimensional data. Effective high-dimensional visualization methods are employed to support visual exploration and inspection of hidden patterns inside the distributions within both spaces. These data abstraction and visualization methods can be easily adapted to various MOEAs as long as they follow the same data protocol, where solutions are represented by vectors in two spaces. In addition, such abstraction facilitates the visualization design by allowing a unified representation across varying numbers of decision and objective variables, which also reduces the learning cost of using the framework.

Despite the intended adaptability from the perspective of MOEAs, certain challenges related to the scalability of ParetoLens should also be addressed, particularly in managing large numbers of solutions and handling excessively high-dimensional data within the visualizations:

\begin{itemize}[leftmargin=*]
    \item When dealing with vast quantities of solutions, such as tens of thousands, issues like over-plotting can occur, particularly in scatterplots where dots represent solutions and in the PCP where lines represent edges. To alleviate over-plotting in scatterplots, semantic zooming has been implemented, allowing users to magnify specific areas and thereby uncovering more details in those areas. Additionally, to further address this challenge, employing advanced sampling techniques~\cite{Yuan2021a} could serve as a feasible strategy. These techniques aim to maintain the distributional integrity of the solution sets while reducing visual clutter.
    \item For the issue of high dimensionality, while t-SNE effectively reduces dimensions for visualization, the PCP faces a bottleneck due to the limited horizontal space available in the interface. The readability of the visualization and the interactions with the PCP on a standard monitor, which is in a pixel resolution of 1920$\times$1080, may gradually decrease when the total number of decision and objective variables exceeds 30. This observation can be partially reflected in the result in Figure~\ref{fig:case1} (D). Such constraint can hinder the clarity and readability of the visualization when displaying high-dimensional results for large-scale optimization~\cite{tian2021esurvey, liu2023survey}. A potential improvement to enhance the PCP's capacity could involve integrating toggles that allow users to display or hide specific axes~\cite{Johansson2015} selectively. This feature would enable users to tailor the visualization to their needs, focusing on relevant dimensions while minimizing visual overload.
\end{itemize}

\subsection{Comparison with Existing Methods}
To provide a clearer understanding of our research within the context of existing literature, a comparison is presented in Table~\ref{tab:comparison}, highlighting the distinctions among selected representative methods based on key characteristics. The ``method type'' column differentiates between the nature of the approach employed by each work --- whether it is a visualization algorithm which typically yields a static image, or a system designed for interactive engagement. Notably, Pymoo~\cite{pymoo}, classified as a computational library, supports visualizing solutions through static charts.

Regarding visualization design and interactivity, the first two algorithms rely on high-dimensional projection techniques as their primary strategy. In contrast, ParetoLens is grounded in the visual analytics paradigm, offering a comprehensive suite of interactive visualization components. These components effectively address various analytical tasks, which sets ParetoLens apart from conventional single-view outputs. While most existing methodologies generate only a single output view, ParetoLens introduces coordinated multi-view support. This design enables dynamic interaction across different visualizations, significantly enhancing visual comprehension and exploration.

\subsection{Limitations and Future Work}
While the current design of ParetoLens offers substantial potential, there are several areas for enhancement. First, the current approach abstracts data visualization from specific application contexts, treating test problem data without reference to real-world scenarios. Although this design aligns with common EMO protocols, it limits the framework's ability to convey the semantic meanings of solutions in particular application domains. Enriching the user experience with domain-specific visualizations could significantly enhance decision-making support. For instance, in the context of mechanical structure design problems~\cite{He2020}, incorporating visualizations that represent the actual structural designs of solutions could improve the user's understanding. One promising future extension is to evolve ParetoLens into a plugin-based system, thereby enabling developers to integrate new visualizations and interactive features tailored to specific application scenarios.

Additionally, enhancing the reporting of details about the evolutionary processes of MOEAs could extend the framework's adaptability. Attributes such as the number of iterations or population statistics across generations could be incorporated into the information panels on the left side of the interface as additional items. This would provide users with deeper insights into the evolutionary processes, further supporting MOEA analyses. In addition, for test problems with specified constraint regions, methods will be explored to represent region information in the high-dimensional projection results.

Future work will also focus on enabling comparisons between multiple solution sets, whether from multiple runs of the same algorithm or from different algorithms. This initiative arises from feedback during expert interviews, where the need for such functionality was highlighted. Comparison in visualization design is complex and requires careful attention to avoid bias and reduce visual clutter~\cite{Gleicher2017}. Future efforts will explore methodologies for effectively visualizing similarities and differences among solution sets, enhancing visual exploration and supporting algorithm comparison.

Another area of focus is the impact of tunable hyperparameters, particularly for the t-SNE and HDBSCAN methods used in generating visual mappings. Currently, the framework relies on default or recommended settings from the supporting libraries\footnote{\url{https://github.com/lmcinnes/umap}}\footnote{\url{https://scikit-learn.org}} . Future research will investigate the effects of hyperparameter adjustments on these methods and assess their influence on visualization outcomes.

Finally, based on expert feedback, there is a demand for exporting intermediate analytical findings as static images. Although ParetoLens excels in interactive visual analytics, there are scenarios where real-time demonstrations may not be feasible. In such cases, the ability to generate static outputs (e.g., printed charts or images) will be crucial for documenting and communicating key insights. Future developments will explore ways to capture snapshots of visualization results, providing customizable options for restyling content in scatterplots and parallel coordinate plots.

\section{Conclusion}
In this paper, a visual analytics framework, \textit{ParetoLens}, is proposed to facilitate the inspection and exploration of multi-objective solution sets generated by multi-objective evolutionary algorithms. The design of ParetoLens is informed by the nested model for visualization research as well as insights from the preliminary study identifying specific challenges and tasks. The framework features a multi-aspect design, enabling comprehensive analysis of data distributions and providing a value-centric perspective for inspecting solution sets alongside their corresponding reference points. The effectiveness and usability of ParetoLens are demonstrated via case studies and expert interviews.

\ifCLASSOPTIONcaptionsoff
  \newpage
\fi



\bibliographystyle{IEEEtran}
\bibliography{references}
\end{document}